\documentclass[10pt,twocolumn,letterpaper]{article}

\usepackage{wacv}
\usepackage{times}
\usepackage{epsfig}
\usepackage{graphicx}
\usepackage{amsmath}
\usepackage{amssymb}
\usepackage{algorithm} 
\usepackage{algpseudocode} 
\usepackage{flexisym}
\usepackage[dvipsnames]{xcolor}
\usepackage{svg}
\usepackage[nodisplayskipstretch]{setspace}
\usepackage{ctable}
\usepackage{array, boldline, makecell, booktabs}

\wacvfinalcopy % *** Uncomment this line for the final submission
\ifwacvfinal
\usepackage[breaklinks=true,bookmarks=false]{hyperref}
\else
\usepackage[pagebackref=true,breaklinks=true,colorlinks,bookmarks=false]{hyperref}
\fi

\begin{document}

%%%%%%%%% TITLE
\title{
MSNet: A Multilevel Instance Segmentation Network for Natural Disaster Damage Assessment in Aerial Videos

}

\author{Xiaoyu Zhu\\
Carnegie Mellon University\\
{\tt\small xiaoyuz3@cs.cmu.edu}\\
\and
Junwei Liang\\
Carnegie Mellon University\\
{\tt\small junweil@cs.cmu.edu}
\and
Alexander Hauptmann\\
Carnegie Mellon University\\
{\tt\small alex@cs.cmu.edu}
}

\begin{figure*}
\vspace{6mm}
\twocolumn[{%
\renewcommand\twocolumn[1][]{#1}%
\maketitle
\begin{center}
\centering
\vspace{-10mm}
   \includegraphics[width=1.0\textwidth, height=4.5cm]{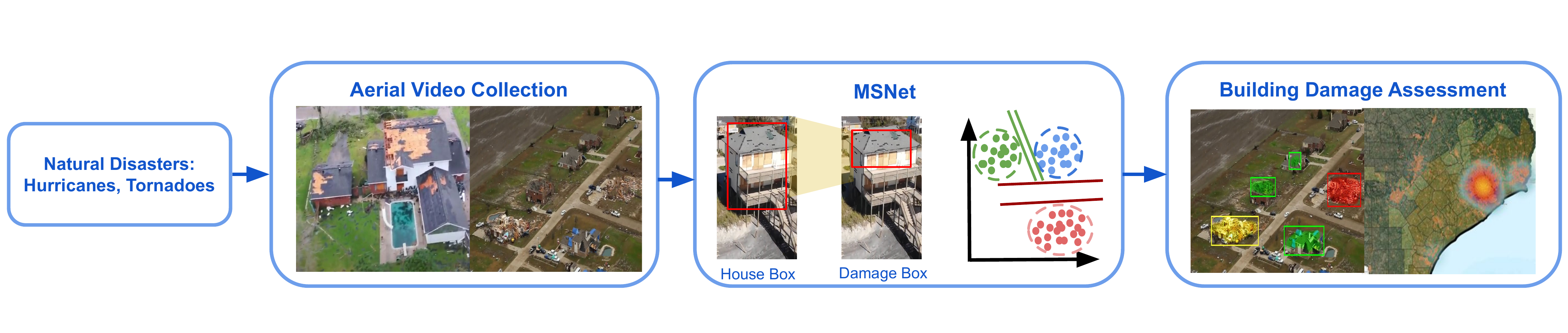}
   \vspace{-7mm}

    \caption{Illustration of the natural disaster damage assessment pipeline. Aftermaths of natural disasters are recorded by drones. Our model is able to detect damage masks and damage scales in different locations. The damage detections along with drones' GPS trajectory could generate a damage assessment location heatmap to aid timely disaster relief efforts.}
 
\label{fig:overview}
\end{center}%
}]
% \label{fig:overview}
\end{figure*}

%%%%%%%%% ABSTRACT
\begin{abstract}
    In this paper, we study the problem of efficiently assessing building damage after natural disasters like hurricanes, floods or fires, through aerial video analysis. We make two main contributions. The first contribution is a new dataset, consisting of user-generated aerial videos from social media with annotations of instance-level building damage masks. This provides the first benchmark for quantitative evaluation of models to assess building damage using aerial videos. The second contribution is a new model, namely MSNet, which contains novel region proposal network designs and an unsupervised score refinement network for confidence score calibration in both bounding box and mask branches. We show that our model achieves state-of-the-art results compared to previous methods in our dataset.\footnote{\url{https://github.com/zgzxy001/MSNET}}

\end{abstract}

%%%%%%%%% BODY TEXT

\section{Introduction}

In recent years, natural disasters have impacted many vulnerable areas around the world. In 2019, there have been ten natural disaster events with damages of more than 1 billion dollars each across the United States \cite{noaa}. Timely response to natural disasters plays a crucial role in disaster relief. However, current damage assessments are mostly based on manual damage detection and documentation, which is slow, expensive and labor-intensive work \cite{fema}. 

With the increasing availability of consumer-grade drones, a large number of aerial videos are recorded and shared across social media~\cite{liang2017temporal}.
After a natural disaster, like a hurricane or a flood, 
people frequently share drone footage of the district, or the authorities could dispatch drones themselves to assess the damage of the area.
These videos could serve as valuable resources for automatic damage assessment. 
Compared with satellite imagery used in previous damage assessment task works \cite{chen2018benchmark,gupta2019xbd,10.1117/1.JRS.11.016042}, drone videos have the advantage of capturing detailed observations of each building from different angles other than just from a top-down perspective.
Valuable structural information of the buildings could be extracted from drone videos for further damage evaluation, \emph{i.e.}, whether the buildings are going to collapse.

Consider the example in Figure~\ref{fig:overview},
there are three challenges for automatic building damage assessment. 
The first is the diversity of buildings, the level of damages and the location of damages. Buildings could include homes, schools, coastal buildings, factories, and other facilities. Some might be slightly damaged, and others might be completely damaged. Some might only have severe damage on the roof.
The second challenge is the detection of small objects and debris. 
The drone videos are usually recorded from a high altitude where many of the damaged parts are only represented by a few dozen pixels (See Section~\ref{sec:dataset}).
The third challenge is the changes of viewpoints as the drone flies over the area.
The damage of a building might only be visible from a certain viewpoint.
This leads to problems like missed detection and inconsistent detections by a single image-based detector.

To overcome the aforementioned challenges, we have collected the first dataset with aerial videos for natural disaster damage assessment.
Our dataset, namely ISBDA (Instance Segmentation in Building Damage Assessment), consists of fine-grained building damage bounding box and mask annotations of different damage levels.
This provides the first quantitative benchmark for evaluating building damage assessment models.
Our second contribution is to propose a new neural network model, \emph{MSNet}, to address the difficulties of accurately detecting damages in buildings with aerial videos.
Our model makes use of the hierarchical relationship between building and damage, and inter-frame spatial consistency of multiple viewpoints to train more robust representations. 
To summarize, our contribution is fourfold:
    \begin{itemize}
    \item We present the first natural disaster building damage assessment dataset, namely ISBDA, using aerial drone videos. It is annotated with fine-grained instance-level building and damage bounding boxes and masks. It provides the first quantitative benchmark for assessing damage assessment in aerial videos.

    \item We propose a novel neural model termed Hierarchical Region Proposal Network (HRPN), which explores the hierarchical spatial relationship among different objects, and thus significantly improving the model performance. 
    \item We propose an unsupervised score refinement model named Score Refinement Network (SRN) based on inter-frame consistency to tackle the challenges of detections using drone videos.
 
    \item We empirically validate our model on the proposed ISBDA dataset for damage assessment, in which our model achieves the best results compared to state-of-the-art object detection models. 
    \end{itemize}

%-------------------------------------------------------------------------

%------------------------------------------------------------------------
\section{Related Work}
\paragraph{Natural Disaster Damage Assessment Datasets.} 
Existing damage assessment dataset can be roughly categorized into two types: ground-level images and satellite imagery. The ground-level images were mostly collected from social media \cite{10.1145/3110025.3110109}. Those datasets only have image-level labels available, because the scene captured by a single ground-level image is highly limited. Besides, due to the lack of geo-tags in social media, ground-level images may not be suitable for large-scale damage assessment. Another disaster data source is satellite imagery based on remote sensing \cite{chen2018benchmark,gupta2019xbd,10.1117/1.JRS.11.016042,rudner2018multimathbf3net,doi:10.1080/01431161.2017.1294780}. However, the main limitation of satellite imagery is that it could not provide detailed damage information due to the long distance to the captured buildings and its limited vertical viewpoint. We are the first to propose a dataset from drone video viewpoints (typically about forty-five degrees) for damage assessment tasks with instance-level damage annotations.

\paragraph{Damage Detection Approaches.}
Current damage detection approaches can be put into three categories. The first category is using supervised machine learning methods which include pixel-based relevant change detection \cite{4509590} and object-based local descriptors \cite{6357329}. 
The second category includes unsupervised methods \cite{5776675,4481231,4060945} that generally refer to outlier detection in scene changes. 
The third category, a recent trend on damage assessment is using semi-supervised approaches \cite{Gueguen_2015_CVPR} aimed at using less human-labeled data and maintaining higher accuracy.  
Other literature also proposed deep learning frameworks such as Convolutional Neural Networks (CNN) \cite{Ahmad2017ConvolutionalNN,10.1145/3110025.3110109} to predict the damage level of each image. However, existing models only worked on building bounding box prediction tasks, which lack specific locations of damaged parts. 

\paragraph{Anchor-based Region Proposal Networks.}
Existing literature on anchor-based region proposal networks mostly adopted dense anchoring scheme, where anchors are sampled densely over the spatial feature space with predefined scales and aspect ratios. The most representative work is Region Proposal Network (RPN) introduced in Faster R-CNN \cite{ren2015faster}, which designed a light fully convolutional network to map sliding windows to a low-dimensional feature space. This framework has been widely adopted in later research \cite{dai2016rfcn,he2017mask}. Some research \cite{yang2018metaanchor}  focused on using meta-learning to dynamically generate anchors from the arbitrary customized prior boxes. Other research works \cite{cai2017cascade,chen2019hybrid,zhang2017singleshot} adopted cascade architecture to regress bounding boxes iteratively for progressive anchor refinement. Some researchers \cite{wang2019region} tried to remove the iteration process by predicting the center of objects of interest. However, there is still a lack of region proposal networks that could utilize spatial hierarchical relationships among objects which could potentially improve detection accuracy. 

\begin{figure*}[!htp]
\centering
\includegraphics[width=1.0\linewidth]{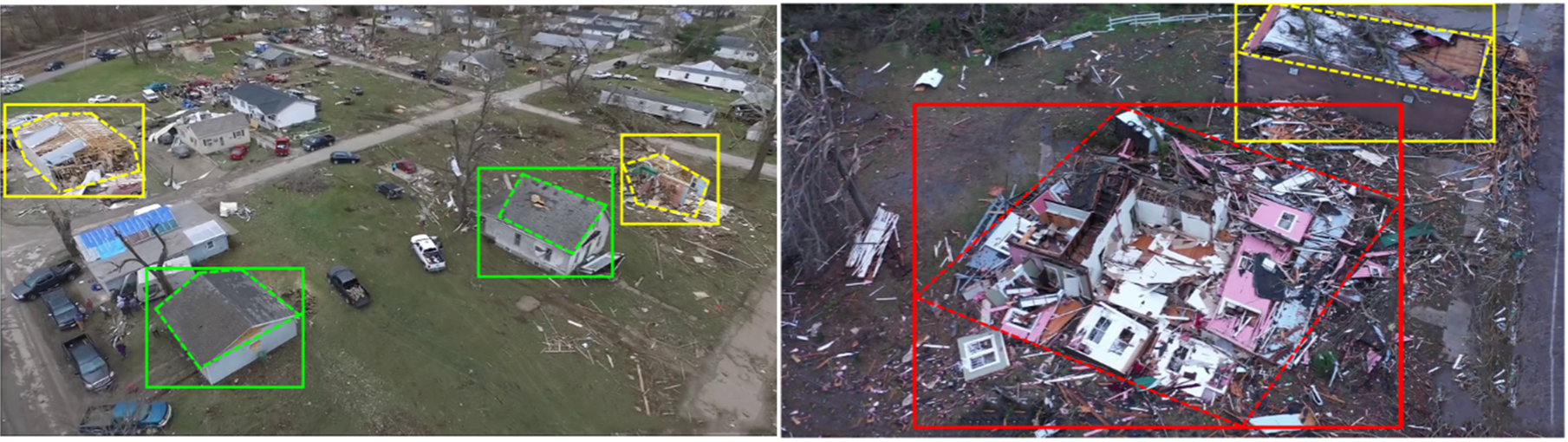}
\caption{Visualization of our ISBDA dataset. The green, yellow and red polygons denote damages in Slight, Severe and Debris levels, respectively. The rectangles composed of solid lines represent damaged building bounding boxes.
The polygons with dotted lines represent segmentation masks of damaged parts.}
\label{fig:anno}
\end{figure*}

\paragraph{Detection Score Refinement.}
Current research in detection score refinement can be categorized into two streams, bounding box score refinement and mask score refinement. In bounding box score correction, most works focused on making modifications on the basis of Non-maximum Suppression (NMS) algorithm, such as Fitness NMS \cite{tychsensmith2017improving} and SoftNMS \cite{bodla2017softnms}. Jiang \etal \cite{jiang2018acquisition} proposed  IoU-Net that directly predicted box IoU, and the predicted IoU was used for the bounding boxes refinement. In terms of score refinement in mask level, Mask Scoring R-CNN \cite{huang2019mask} was proposed by adding a MaskIoU head to regress the IoU between the predicted mask and its ground truth mask. One limitation of this approach is that it can only refine the mask scores, which nearly has no impact on the bounding box branch. Our proposed score refinement algorithm based on inter-frame consistency is able to achieve consistent improvement in both bounding box and mask branches.

\section{The ISBDA Dataset}
\label{sec:dataset}

\subsection{Data Collection}
In order to fully assess building damages in different scenarios and locations, we have collected ten videos from social media platforms, which recorded severe hurricane and tornado disaster aftermaths in recent years. Specifically, the aerial videos were recorded after Hurricane Harvey in 2017, Hurricane Micheal and Hurricane Florence in 2018 and other three tornadoes (EF-2 or EF-3) in 2017, 2018 and 2019, respectively. 
The affected areas recorded in the videos include Florida, Missouri, Illinois, Texas, Alabama and North Carolina in the United States. The total length of the collected videos is about 84 minutes.

To get individual frames, we first obtain video clips from the ten videos that: (1) do not have apparent camera rotations; and (2) fly with moderate and stable speed. To further improve the annotation efficiency and cover different scenarios, we extract one frame out of every ten frames from these video clips. Overall, we have collected 1,030 frames for instance-level building and damage annotation.

One important problem is to define damage scale and corresponding standards which can cover various types of damages in different scenes. Following the damage assessment practice, Joint Damage Scale \cite{gupta2019xbd}, we divide building damages into three levels: Slight, Severe and Debris. Slight refers to visible cracks or appearance damages. Severe refers to partial wall or roof collapse, which are apparent structural damages. Debris refers to completely collapsed buildings. 

\subsection{Hierarchical Instance-level Annotation}
To provide fine-grained localization information of individual damages, we formulate the damage assessment task as an instance segmentation problem. We annotate both the polygons of damaged buildings and the specific damaged parts of the buildings. 
In order to explore the hierarchical relationships between building and damaged part instances (\emph{i.e.}, specific damaged parts are within corresponding damaged building boxes), we also include the mappings between each damaged part ID and its corresponding damaged building ID. The dataset is annotated by three experienced annotators, and one pass of verification is performed for each annotation to ensure accuracy.

\subsection{Dataset Statistics}

Overall, 1,030 images sampled from 10 videos are annotated with instance-level building masks and damaged part masks. 
The dataset has 2,961 damaged part instances which are divided into three levels: Slight, Severe, and Debris. 
Following Microsoft COCO's \cite{lin2014microsoft} size definition, we calculate the number of damaged part instances in different sizes for each damage scale, shown in Table~\ref{table:stas_scale}.

\begin{table}[!htp]
\small 
\centering
\begin{tabular}{c|ccc|c}
Damage Scale & Small & Medium & Large & Total\\
\hline
Slight & 204 & 1169 & 746 & 2119\\
Severe & - & 120 & 440 & 560\\
Debris & - &  54 & 228 & 282 \\
\end{tabular}
\vspace{2mm}
\caption{Distribution of annotation sizes. Small: area less than $32\times32$; Medium: area greater than $32\times32$ and less than $96\times96$; Large: area greater than $96\times96$. Area is measured as the number of pixels in the segmentation mask.}
\label{table:stas_scale}
\end{table}

\begin{figure}[!htp]
% \centering
\includegraphics[width=1.0\linewidth]{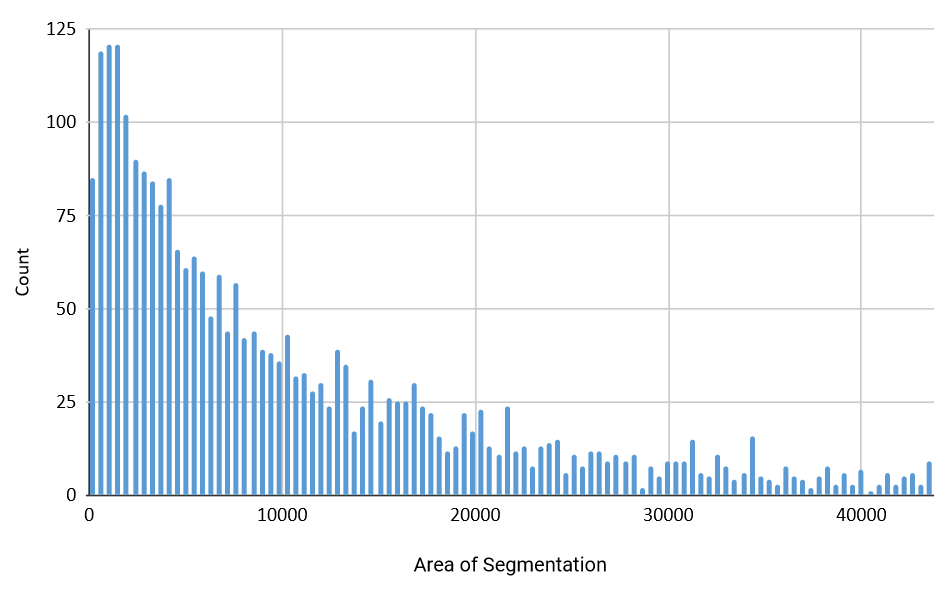}
\caption{The distribution of the area of damage segmentation in our ISBDA dataset. We only show the distribution of areas below 90th percentile of the whole dataset for better visualization purpose. Area is measured as the number of pixels in the segmentation mask.}
\label{fig:distribution}
\end{figure}

We also analyze the distribution of the area of damage segmentation in the ISBDA dataset, shown in Figure~\ref{fig:distribution}. We observe that the majority of the damage segmentation are relatively small. Visualization of the ISBDA dataset and annotations is shown in Figure~\ref{fig:anno}.

\section{Method}

\begin{figure*}[!htp]
\centering
\includegraphics[width=1.0\linewidth, height=7cm]{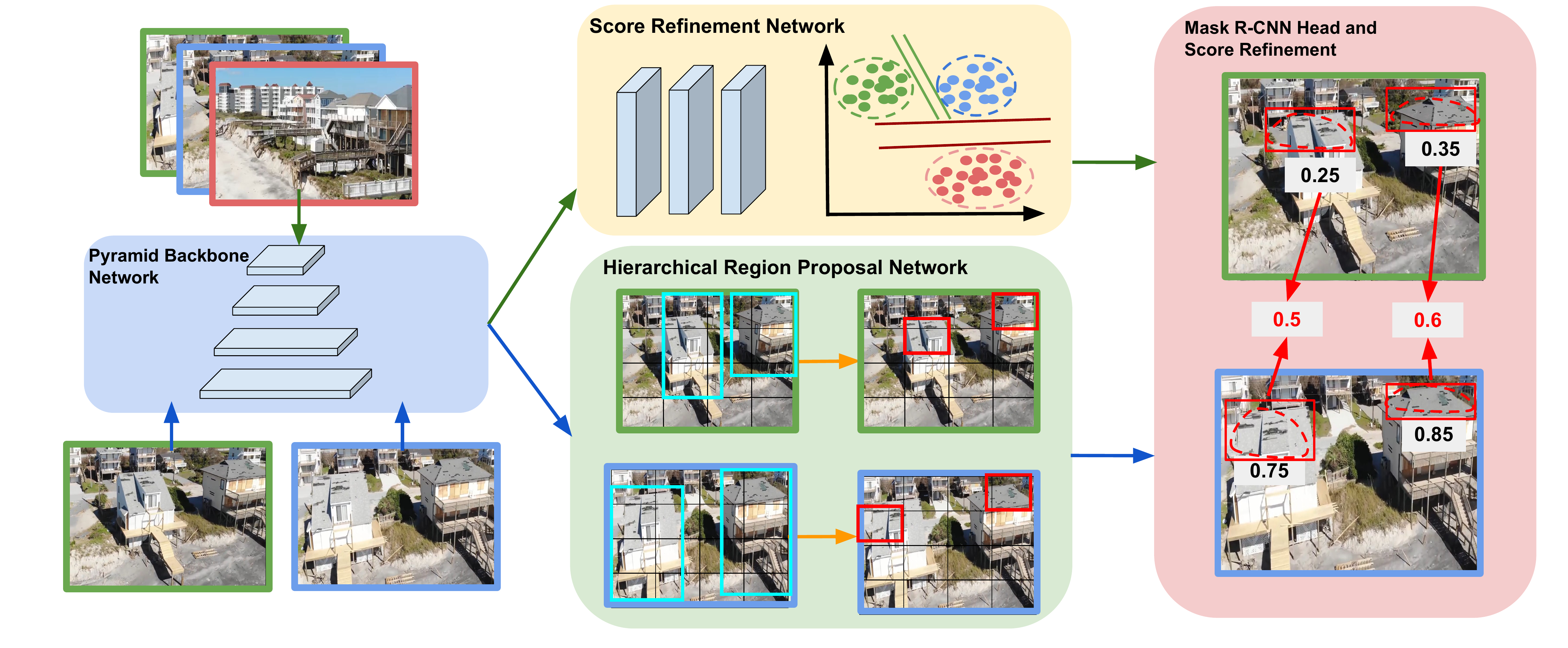}
\caption{Network architecture of \emph{MSNet}. The left part contains a pyramid backbone network to extract features in multi-scale levels. The backbone network is shared in the two neural network's training. The first neural network (Bottom) is for generating instance segmentation results. Specifically, for each image, Hierarchical Region Proposal Network takes the encoded features to generate proposals for damaged buildings. The building proposals are used to give supervision on damage proposals generation (Yellow Arrow). The second branch (Top) is for the training of Score Refinement Network. The adjacent frames (images with green and blue edges) along with one negative sample (image with red edges) are firstly fed into the Pyramid Backbone Network, then Score Refinement Network is trained with the proposed Multi-scale Consistency Loss to learn feature similarity. These two branches are joined at the end, where Mask R-CNN Head generates bounding box and mask predictions. Finally, the score refinement algorithm is performed to calibrate the confidence scores.
} 
\label{fig:is}
\end{figure*}

\subsection{Overview}
To provide fine-grained localization information, similar to some of the existing works \cite{gupta2019xbd}, we formulate the damage assessment task as an instance segmentation problem. Moreover, our model will predict damage-level instance masks instead of building-level, which is a more challenging task due to the high damage variance and small damaged area. We propose a new model named \emph{MSNet} in order to learn more robust representations in different scenarios with different viewpoints. It includes two types of supervision: supervision of building bounding boxes for low-level damage anchor sampling and mask segmentation; and supervision of temporal and spatial relationships between adjacent video frames. In summary, it has the following key components: \\
\noindent\textbf{Pyramid Backbone Network} 
uses ResNet-50 based Feature Pyramid Network (FPN) \cite{lin2016feature} to extract spatial features of input images. \\
\noindent\textbf{Hierarchical Region Proposal Network} 
first generates high-level building proposals and then uses them to supervise low-level anchor sampling and damage proposals generation. \\
\noindent\textbf{Score Refinement Network} is proposed to calibrate the confidence scores of instances in adjacent frames which share common appearance features but have confidence score variances. \\
\noindent\textbf{Mask R-CNN Head} includes the R-CNN head for bounding box and class prediction, and the Mask head for mask prediction \cite{he2017mask}. \\

In the rest of this section, we will introduce the above components and the learning objectives in details. 

\subsection{Hierarchical Region Proposal Network}
\label{sub:hrpn}
Traditional Region Proposal Network (RPN) treats all objects in the same spatial level, and uniformly generates dense anchors over the feature space. If we adopt a conventional RPN scheme and train the RPN with building and damage proposals simultaneously, the hierarchical relationship between buildings and damaged parts will not be utilized. Therefore, we propose a new model, termed Hierarchical Region Proposal Network (HRPN), 
to address the aforementioned problems.

In HRPN, there are two RPNs sharing the same backbone network: a high-level RPN and a low-level RPN. The high-level RPN is trained with damaged building boxes with binary labels indicating whether the proposal is a damaged building or not. The low-level RPN utilizes building proposal outputs from the high-level RPN for anchor sampling. We sample anchors based on one of the two metrics: Intersection over Union (IoU) and Inner Intersection (II) between high-level region proposals and low-level anchors. For each low-level low-level (damage) anchor $A_{\widetilde{a}}$, we define its sampling score as:

{\footnotesize
\begin{equation}
    \text{S}_{IoU}(A_{\widetilde{a}}, A_p) = \max_{A_p \in {P}} \frac{A_{\widetilde{a}} \bigcap A_p}{A_{\widetilde{a}} \bigcup A_p}
\end{equation}}

{\footnotesize
\begin{equation}
    \text{S}_{II}(A_{\widetilde{a}}, A_p) = \max_{A_p \in {P}} \frac{A_{\widetilde{a}} \bigcap A_p}{A_{\widetilde{a}}}
\end{equation}}

where $P$ is a set of high-level (building) region proposals. For each anchor, we compute its sampling score and only keep anchors with scores larger than a certain threshold $S$. Then the sampled anchors are used for damage proposals generation. 

\subsection{Score Refinement Network}
\label{sec:mlc}

In previous works~\cite{bolya2019yolact}, the confidence scores are determined by single-frame detection, while correspondence between two adjacent frames is not utilized. We propose a score refinement model based on inter-frame temporal and spatial correspondence termed Score Refinement Network (SRN). The input of the model is randomly generated triplets and each triplet is composed of one frame and its adjacent frame as a positive frame and another random frame as a negative frame. 
By incorporating multi-scale features from the FPN backbone, we design a multi-scale consistency loss to force SRN to learn feature representations such that one sample's distance to its positive sample is closer than its distance to the negative one.  
We aim to refine the scores of instances in adjacent frames which share common appearance features but have confidence score variances.

Inspired by \cite{wang2015unsupervised}, we use patch mining to build triplets and each is composed of one sample $P_i$, its relative adjacent frame $P_i^{+}$ and its random sample $P_i^{-}$. 
The triplets are sampled based on the fact that the average drone speed is 50 mph and thus the frame variances within half seconds are small.
Therefore, given a frame $x_t$ at time $t$ and the video frame rate $r$, the positive sample is defined as the frame in range \([x_t-0.5r, x_t+
0.5r]\). The negative sample is defined as the frame in range \([0, x_t-10r]\) $\bigcup$ \([x_t+{10r}, T]\). $T$ is the maximum frame number of the video.

Multi-scale features usually demonstrate significant performance improvement in object detection tasks \cite{he2017mask,lin2016feature}. Therefore, we propose Multi-scale Consistency Loss (MCL) which makes use of multi-scale feature maps. For two image patches $X_i$, $X_j$, we firstly obtain the feature maps of each image from the last four layers of the FPN backbone, namely $P_{ik}$, $P_{jk}$, where $k$ $\in$ [1, 2, 3, 4]. These feature maps are used as input to SRN. For an input feature $P$, we can obtain its feature from the last SRN layer as $f(P)$, where $f$ is a feature encoder which is composed of three fully connected layers. Then, we propose a spatial-wise similarity metric of two feature maps $P_{ik}$, $P_{jk}$ in FPN level $k$ using:

{\footnotesize
\begin{eqnarray}\label{eq:cos_dis}
Sim(P_{ik}, P_{jk}) = \sum_{w=0}^{W}\sum_{h=0}^{H} \frac{f(P_{ik}^{wh}) \cdot f(P_{jk}^{wh})} {\|f(P_{ik}^{wh})\| \|f(P_{jk}^{wh})\| }
\label{equ:sim}
\end{eqnarray}
}

{\footnotesize
\begin{eqnarray}\label{eq:cos_dis}
D(P_{ik}, P_{jk}) = 1 - Sim(P_{ik}, P_{jk})
\end{eqnarray}
}

Given a set of triplets and each triplet is denoted as ($X$, $X^{+}$, $X^{-}$) , we aim to train SRN which can learn feature representations such that $D(X, X^{-}) > D(X, X^{+})$ using the Multi-scale Consistency Loss (MCL):

{\footnotesize
\begin{eqnarray}\label{eq:cos_loss}
\mathcal{L}_{mcl}(X, X^{+}, X^{-}) = \sum_{i=1}^{L} \max\{ 0,  D(X_i, X_i^{+}) - D(X_i, X_i^{-}) + m \}
\end{eqnarray}
\label{equ:mcl}
}

where $m$ is a margin constraint parameter, and $L$ is the number of multi-scale layers.

\subsection{Training}

In this section, we provide detailed descriptions of the training procedure. The first part of the loss function is the HRPN loss, which is defined as:

\vspace{3mm}
{\footnotesize
\begin{equation}\label{eq:loss}
\begin{aligned}
\mathcal{L}_{hrpn} = \mathcal{L}_{rpn}^{h} + \mathcal{L}_{rpn}^{l}.
\end{aligned}
\end{equation}
}

Here, $\mathcal{L}_{rpn}^{h}$ and $\mathcal{L}_{rpn}^{l}$ represent the loss of high-level RPN and low-level RPN, respectively. The low-level RPN conducts anchor sampling and proposal generation under the supervision of high-level RPN. As described in Section~\ref{sub:hrpn}, the losses of damage proposals which are filtered out under the supervision of high-level building proposals are not computed in the HRPN loss. The definition of RPN loss follows \cite{ren2015faster}. $\mathcal{L}_{cls}$, $\mathcal{L}_{box}$, and $\mathcal{L}_{mask}$ follow the definitions in \cite{he2017mask}. $\mathcal{L}_{mcl}$ is computed using Equation~\ref{equ:mcl}.

The final multi-task loss of our proposed approach is calculated using:

{\footnotesize
\begin{equation}\label{eq:loss}
\begin{aligned}
\mathcal{L} = \mathcal{L}_{hrpn} + \mathcal{L}_{cls} + \mathcal{L}_{box} + \mathcal{L}_{mask} + \mathcal{L}_{mcl}.
\end{aligned}
\end{equation}
}

The HRPN and Mask R-CNN Head can be trained end-to-end together with SRN. However, in that case, the model training and inference would be heavy due to the multi-scale feature similarity calculation. Therefore, we only calibrate confidence scores of the model which has the best instance segmentation performance.

\subsection{Inference}
\label{sec:inf}

In test time, we use HRPN to generate building region proposals. Then the building proposals are used as supervision for damage anchor sampling and proposal generation, as described in Section~\ref{sub:hrpn}. In the second stage, the model extracts features using RoIAlign for each damage proposal and performs proposal classification, bounding box regression and mask prediction.

During the inference of SRN, given two adjacent frames $P$ and $Q$, we firstly extract the last four layers from the Pyramid Backbone Network for each frame. The four layers are used as input for SRN described in Section~\ref{sec:mlc} to extract similarity feature maps. Then we use RoIAlign to align the extracted features with each bounding box. For each prediction (including bounding box and mask) in frame $P$, we calculate its similarity score with each prediction in frame $Q$, using equation~\ref{equ:sim} with the aligned feature maps as input. Then we can obtain the prediction in frame $Q$ that has the highest similarity score with it. The average of these two confidence scores is used as their final scores. Note that we only refine confidence scores that fall within the range of [$C_0$, $C_1$].

\begin{table*}[t]
\centering
 \vspace{3mm}
\begin{tabular}[!ht]{
p{5cm}|p{1cm}|p{1cm}|p{1cm}|p{1cm}|p{1cm}|p{1cm} }

Method & AP &AP$_{25}$&AP$_{50}$&AP$^{bb}$&AP$_{25}^{bb}$&AP$_{50}^{bb}$\\
\Xhline{1pt}
  PolarMask+Damage   & 22.3    &29.1&   15.4& 24.4  &29.6&   18.2\\
 Mask R-CNN+Damage   & 34.4    &40.6&   26.9& 35.9  &40.9 &  29.4\\
 Mask R-CNN+Building+Damage &   32.2 & 39.5   &23.3& 34.0    &40.3&   25.7\\
 \textbf{Ours} &\textbf{37.2}& \textbf{44.2}&  \textbf{28.8}& \textbf{38.7}    &\textbf{44.4}&  \textbf{31.5}\\
%  \hline
\end{tabular}
\vspace{2mm}
\caption{Cross scene evaluation results. We report detection and instance segmentation results. AP denotes instance segmentation results and AP$^{bb}$ denotes bounding box detection results. In the results area, rows 1 and row 2 use the PolarMask and Mask R-CNN frameworks with only damage masks as input; row 3 uses Mask R-CNN co-trained with damaged buildings and damages as the baseline model. The results show that our proposed method gains significant improvements compared to state-of-the-art models.}
\label{table:sota}
\end{table*}

\section{Experiments}
\begin{figure*}[!htp]
\centering
% \includesvg[width=1.0\linewidth]{fig/model_v1}
\includegraphics[width=0.95\linewidth,height=10cm]{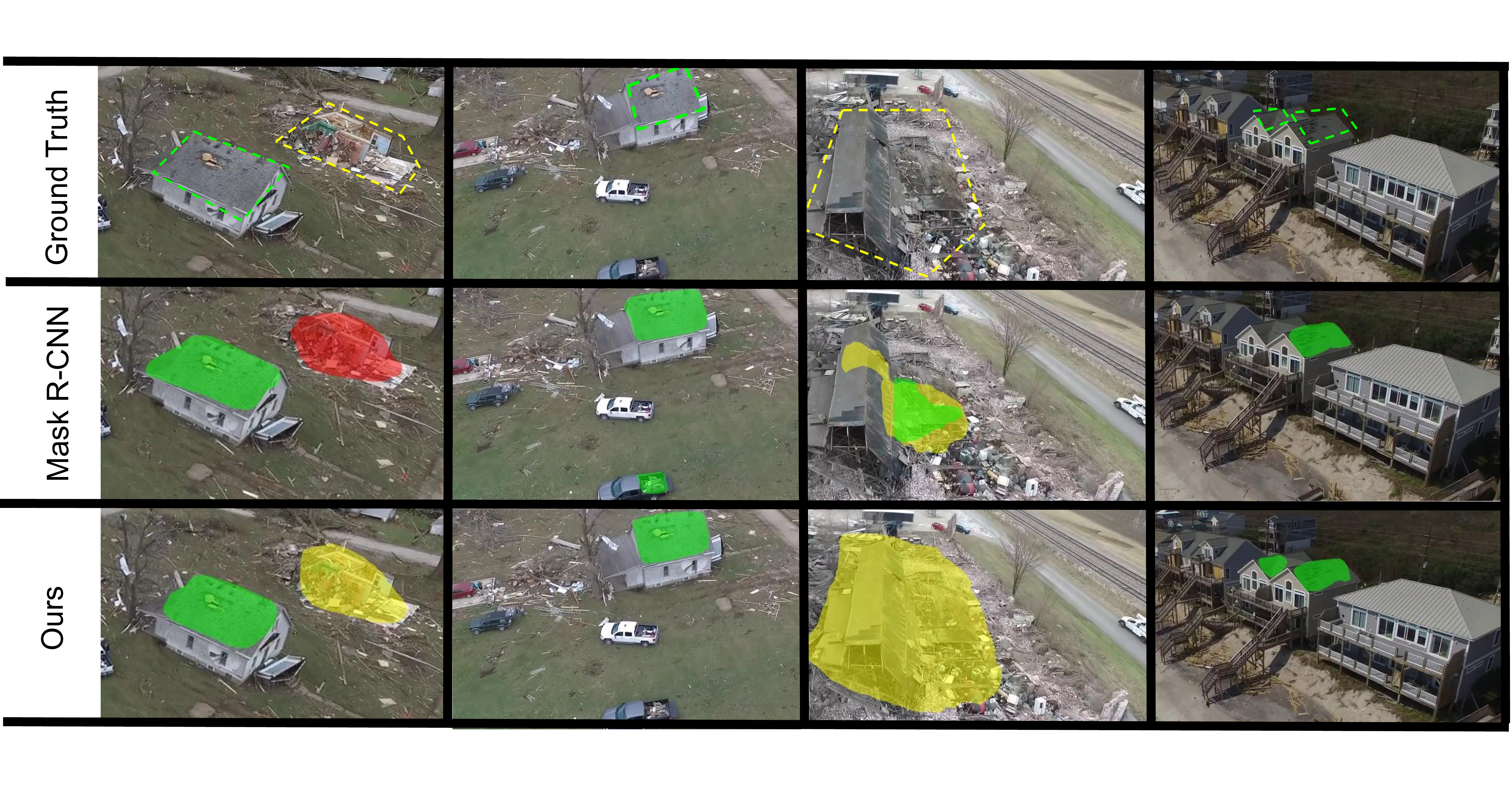}
% ,height=15cm,keepaspectratio
\caption{Visualization of the predicted damage segmentation. This figure demonstrates that our proposed model can alleviate the following errors: (1) label misclassification (first column, left to right); (2) false positive segmentation in the complex scenario with cars and buildings (second column); (3) incompleted masks in noisy video scenario (third column); and (4) missed masks (fourth column).}

\label{fig:sota_vis}
\end{figure*}

In this section, we compare our \emph{MSNet} model with state-of-the-art baselines on the proposed ISBDA dataset.
We randomly split the dataset into subsets with no overlapping scenes.
We train our model using 80\% of the dataset, and test on the rest 20\% dataset.
We repeat the split and experiments 3 times and report the results in Table~\ref{table:sota}. 
The final reported results are the average over the evaluation results of all splits.

We report the standard COCO instance segmentation metric~\cite{lin2014microsoft} including AP (averaged over all IoU thresholds), AP@0.25, AP@0.5, and AP$_S$, AP$_M$, AP$_L$ (AP at different scales). Unless noted, AP is evaluating using mask IoU.

\subsection{Implementation Details}

We compare our model with two recent state-of-the-art instance segmentation models, PolarMask \cite{xie2019polarmask} and Mask R-CNN \cite{he2017mask}. All models use ResNet-50 based FPN as a backbone network. 
We train all the networks for 100 epochs, with a starting learning rate of 0.003 then we decrease it to 0.001 after 10 epochs. Mini-batch SGD is used as the optimizer with batch size equals 8. We initialize all the backbone networks with the weights pre-trained on COCO \cite{lin2014microsoft}. The input images are resized to have the shorter side being 800 and the longer side less or equal to 1333. For testing, an NMS with threshold 0.5 is used and top 100 detections are retained for each image.

For the score refinement procedure, SRN is trained using hard negative mining. We firstly generate 1,000 (X, X$^{+}$) pairs from different videos, and randomly extract 5 negative samples for each (X, X$^{+}$) pair as described in Section~\ref{sec:mlc}. We calculate the loss of 5 negative samples, and choose the top $K$ ones with the highest losses as in \cite{wang2015unsupervised} to optimize. 
For the experiments, we use $K=1$. Adam optimizer~\cite{kingma2014adam} is used for network training with learning rate 0.001, and each batch is composed of one (X, X$^{+}$) pair and 5 negative samples. For testing, we choose $C_0$ = 0.2, and $C_1$ = 0.7 for the range described in Section~\ref{sec:inf}.

\subsection{Comparison to state-of-the-art}
\begin{table*}[t]
\centering
 \vspace{3mm}
\begin{tabular}[!ht]{
p{3.7cm}|p{1.5cm} p{1.5cm} p{1.5cm}|p{1.5cm} p{1.5cm} p{1.5cm}  }
Model & AP &AP$_{25}$&AP$_{50}$&AP$^{bb}$&AP$_{25}^{bb}$&AP$_{50}^{bb}$\\
\Xhline{1pt}
  Baseline   & 35.0    &41.9&   27.8& 36.8  &42.9&   29.9\\
 \hline
  Baseline + HRPN   & 39.3{\color{ForestGreen}\footnotesize\textbf { (+4.3)}}   &46.6{\color{ForestGreen}\footnotesize\textbf { (+4.7)}}&   31.0{\color{ForestGreen}\footnotesize\textbf { (+3.2)}}& 
  41.4{\color{ForestGreen}\footnotesize\textbf { (+4.6)}}  &47.1{\color{ForestGreen}\footnotesize\textbf { (+4.2)}} &  33.7{\color{ForestGreen}\footnotesize\textbf { (+3.8)}}\\
  Baseline + HRPN + SRN  & 40.0{\color{ForestGreen}\footnotesize\textbf { (+5.0)}} 
  &47.7{\color{ForestGreen}\footnotesize\textbf { (+5.8)}} &   
  31.3{\color{ForestGreen}\footnotesize\textbf { (+3.5)}} & 
  42.1{\color{ForestGreen}\footnotesize\textbf { (+5.3)}}   &
  48.1{\color{ForestGreen}\footnotesize\textbf { (+5.2)}}  &  
  33.9{\color{ForestGreen}\footnotesize\textbf { (+4.0)}} \\

\end{tabular}
\vspace{2mm}
\caption{Effect of HRPN and SRN. We use Mask R-CNN co-trained with building and damage instances as the baseline model. The results show that HRPN component gains significant improvement by 4.3\% AP compared with the baseline model. Combined with HRPN, the SRN component also gets consistent improvement in both bounding box and mask branches.}
\label{table:ind}
\end{table*}

\paragraph{Baseline methods.}
We compare our method with state-of-the-art models and their variants customized for the damage instance segmentation problem. PolarMask \cite{xie2019polarmask} is a single shot instance segmentation model with damage masks as input only. Mask R-CNN \cite{he2017mask} is one of the state-of-the-art instance segmentation models. Two variants of Mask R-CNN are used as baselines: (1) Mask R-CNN with damage bounding boxes and masks as input; and (2) Mask R-CNN co-trained with damaged buildings and damages. Damaged building bounding boxes are used for RPN and R-CNN head training, and damage masks are used for the training of Mask head.  

\paragraph{Quantitative results.}
Table~\ref{table:sota} lists the damage instance segmentation results. 
Compared with PolarMask, our model is able to obtain significant improvement, \emph{e.g.}, an absolute increment of 14.9\% mask AP. 
For the Mask R-CNN baselines, we observe that Mask R-CNN trained with damage masks could be confused by the high variance of damage masks in different locations and scenarios. When the Mask R-CNN model is trained with building boxes and damage masks, the errors in building detection will impact the damage detection in the second stage. Also, the model could not precisely predict the damage masks from large building bounding boxes. Our proposed model utilizes the hierarchical nature of the damaged buildings and damaged parts, and outperforms the baseline with 5.0\% AP in the segmentation branch and 4.7\% AP in the bounding box branch. 

\paragraph{Qualitative analysis.}
We qualitatively demonstrate the advantages of our model in  Figure~\ref{fig:sota_vis}, showing that our proposed model can alleviate the following errors: (1) label misclassification (first column); (2) false positive segmentation in the complex scenario with cars and buildings (second column); (3) incompleted masks in noisy video scenario (third column); and (4) missed masks (fourth column). Thanks to the HRPN module and the inter-frame supervision, our model is able to generate accurate and robust detections even in very noisy scenarios like the third column of Figure~\ref{fig:sota_vis}. 
\subsection{Ablation Study}

We evaluate our method on the ISBDA dataset. We use ResNet-50 FPN as a backbone network for ablation study. All experiments in this section are performed on one split.

\begin{figure}[!htp]
\centering
\includegraphics[scale=0.65]{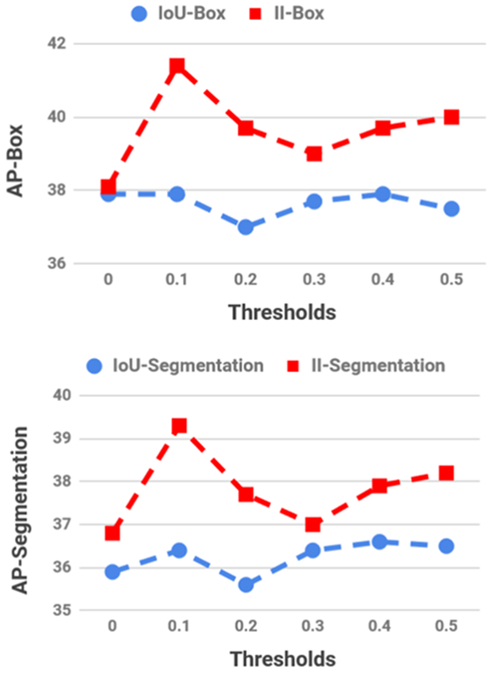}
\caption{mAP of bounding box and segmentation using different IoU and II thresholds. The blue and red lines denote IoU and II metrics, respectively.}
\label{fig:box_iou_ii}
\end{figure}

\paragraph{Different IoU and II thresholds.}
In Figure~\ref{fig:box_iou_ii}, we compare the effects of different thresholds for IoU and II on the model performance using equations in Section~\ref{sub:hrpn}. We train our model with IoU and II from 0.0 to 0.5 in steps of 0.1. For the model with IoU as metrics, the model gets the best performance when IoU equals 0.4. For the model with II as metrics, the model achieves the best performance when it equals 0.1.

\paragraph{Choices of IoU and II metrics.}

In Table~\ref{label:sm}, we report the best performance model among different IoU and II thresholds, respectively, where IoU equals 0.4 and II equals 0.1. We observe that II metric gains 2.7\% AP improvement compared with IoU metric. By analyzing the AP in different sizes, we find that the small objects get the most significant improvement for 7.1\% absolute value. This is probably because in IoU calculation, small damage anchors only occupy a small portion of its union with a large building bounding box. Therefore, small damage instances may not be well detected. On the other hand, II could properly handle such cases as it performs anchor sampling by calculating the intersection within the damage anchors.

\begin{table}
\vspace{3mm}
  \centering
{\begin{tabular}{ p{0.5cm}|p{0.7cm}|p{0.7cm}|p{0.7cm}|p{0.7cm}|p{0.7cm}|p{0.7cm}  }
%  \hline
%  \multicolumn{7}{||c||}{ Different Metrics } \\
M & AP &AP$_{25}$&AP$_{50}$&AP$_{S}$&AP$_{M}$&AP$_{L}$\\
\Xhline{1pt}
  IoU   & 36.6    &42.5&   30.1& 47.4  &41.1&   38.6\\
  II   & 39.3    &46.6&   31.0& 54.5  &38.0 &  42.0\\
%  \hline
\end{tabular}}
\vspace{2mm}
\caption{Results of different anchor sampling metrics.}
\label{label:sm}
\end{table}

\paragraph{Effect of HRPN and SRN.}
In Table~\ref{table:ind}, we experiment with the effect of HRPN and SRN. We observe that the HRPN component gains significant improvement by 4.3\% AP  compared with the baseline model. The SRN component further improves the model performance in both bounding box and mask branches.

%-------------------------------------------------------------------------

%-------------------------------------------------------------------------
\section{Conclusion}
In this paper, we investigate the problem of conducting damage assessment using user-generated aerial video data. We provide the first benchmark, namely ISBDA, for quantitative evaluation for models to assess building damage in aerial videos. Also, our proposed \emph{MSNet} is able to explore the hierarchical spatial relationship among different objects and calibrate confidence scores to improve the model performance in both bounding box and mask branches. We empirically validate our model on the proposed ISBDA dataset, in which our model achieves the best results compared to state-of-the-art object detection models. We believe our dataset, together with our models, will facilitate future research in remote sensing and damage assessment for better and faster natural disaster relief.

\paragraph{Acknowledgements} This research was supported by the financial assistance award 60NANB17D156 from NIST. The views and conclusions contained herein are those of the authors and should not be interpreted as necessarily representing the official policies or endorsements, either expressed or implied, of NIST or the U.S. Government.

{\small
\bibliographystyle{ieee_fullname}
\bibliography{egbib}
}
\end{document}